\documentclass[11pt,a4paper]{article}
\usepackage{authblk}
\usepackage[hyperref]{emnlp-ijcnlp-2019}
\usepackage{times}
\usepackage{latexsym}
\usepackage{url}
\usepackage{algorithm}
\usepackage{tikz}
\usepackage{algpseudocode}
\usepackage{amssymb}
\usepackage{amsmath}
\usepackage{amsthm}
\usepackage{algorithmicx}
\usepackage{verbatim}
\usepackage{bm}
\usepackage{dsfont}
\usepackage{mathtools}

\usepackage[T1]{fontenc}

\usepackage{aecompl}
\usepackage{array}
\usepackage{enumitem}
\usepackage{adjustbox}
\usepackage{subfigure}
\usepackage{tabulary}

\aclfinalcopy 

\newcommand\MN{AdaMRC }

\newcommand{\R}{\mathbb{R}}

\title{Adversarial Domain Adaptation for Machine Reading Comprehension}
\author{\textbf{Huazheng Wang}$^{1}$\thanks{~~Most of this work was done when the first author was an intern at Microsoft Dynamics 365 AI Research.}, \textbf{Zhe Gan}$^{2}$, \textbf{Xiaodong Liu}$^{3}$, \textbf{Jingjing Liu}$^{2}$, \textbf{Jianfeng Gao}$^{3}$, \textbf{Hongning Wang}$^{1}$ \\
$^{1}$University of Virginia,\quad$^{2}$Microsoft Dynamics 365 AI Research, \quad$^{3}$Microsoft Research \\
  \small{\texttt{\{hw7ww,hw5x\}@virginia.edu}, \quad \texttt{\{zhe.gan,xiaodl,jingjl,jfgao\}@microsoft.com }} }
\date{}

\begin{document}
\maketitle
\begin{abstract}
In this paper, we focus on unsupervised domain adaptation for Machine Reading Comprehension (MRC), where the source domain has a large amount of labeled data, while only unlabeled passages are available in the target domain. 
To this end, we propose an Adversarial Domain Adaptation framework (AdaMRC), where ($i$) pseudo questions are first generated for unlabeled passages in the target domain, and then ($ii$) a domain classifier is incorporated into an MRC model to predict which domain a given passage-question pair comes from. The classifier and the passage-question encoder are jointly trained using adversarial learning to enforce domain-invariant representation learning. 
Comprehensive evaluations demonstrate that our approach ($i$) is generalizable to different MRC models and datasets, ($ii$) can be combined with pre-trained large-scale language models (such as ELMo and BERT), and ($iii$) can be extended to semi-supervised learning.
%
%
%
\end{abstract}

\section{Introduction}
Recently, many neural network models have been developed for Machine Reading Comprehension (MRC), with performance comparable to human in specific settings~\cite{gaosurvey}.
However, most state-of-the-art models~\cite{seo2016bidirectional, P18-1157, yu2018qanet} 
rely on large amount of human-annotated in-domain data to achieve the desired performance. Although there exists a number of large-scale MRC datasets~\cite{rajpurkar2016squad,trischler2016newsqa,nguyen2016ms,zhang2018record}, collecting such high-quality datasets is expensive and time-consuming, which hinders real-world applications for domain-specific MRC.  

Therefore, the ability to transfer an MRC model trained in a high-resource domain to other low-resource domains is critical for scalable MRC.
While it is difficult to collect annotated question-answer pairs in a new domain, it is generally feasible to obtain a large amount of unlabeled text in a given domain. In this work, we focus on adapting an MRC model trained in a source domain to other new domains, where only unlabeled passages are available.

\begin{figure}[t]
\centering
\begin{tabular}{cc}
\includegraphics[width=0.47\columnwidth]{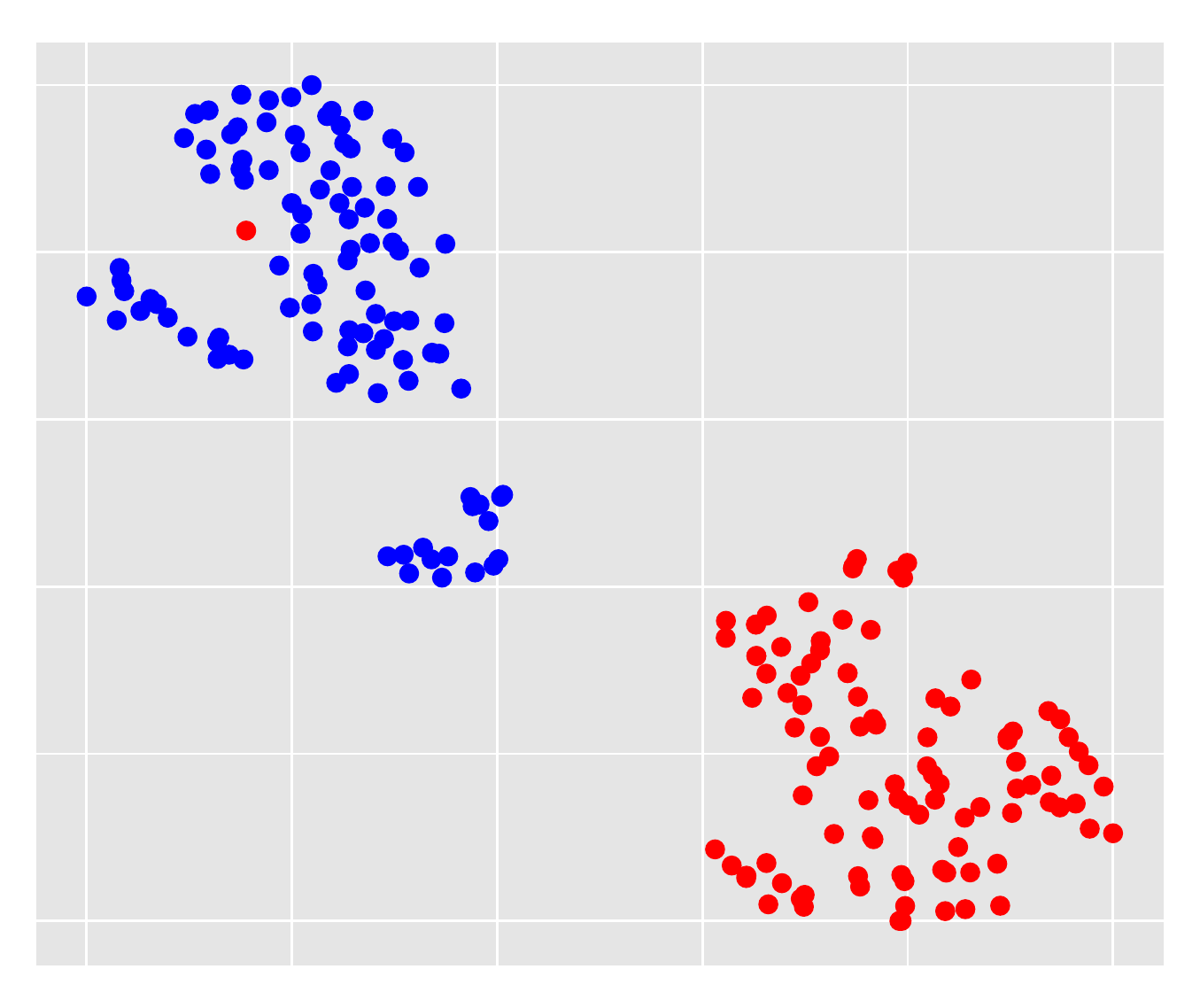} & 
\includegraphics[width=0.47\columnwidth]{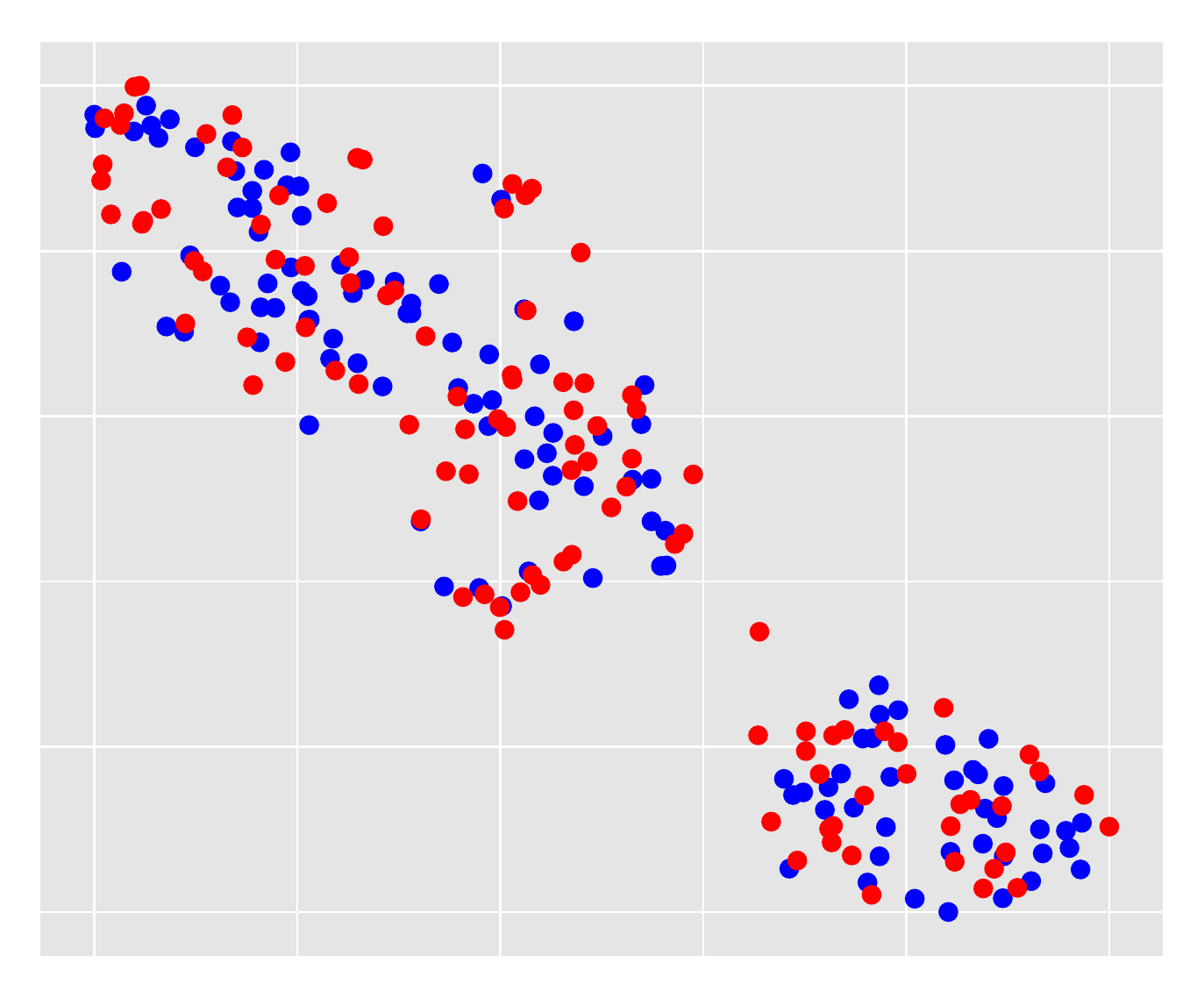}  \\
\small{(a) SynNet} & \small{(b) AdaMRC}
\vspace{-2mm}
\end{tabular}
\caption{t-SNE plot of encoded feature representations from (a) SynNet~\cite{golub2017two} and (b) the proposed AdaMRC. We sampled 100 data points, each from the development set of the source and the target domains. Blue: SQuAD. Red: NewsQA.}
\label{fig:tsne}
\end{figure}

This domain adaptation issue has been a main challenge in MRC research, and the only existing work that investigated this was the two-stage synthesis network (SynNet) proposed in \citet{golub2017two}. Specifically, SynNet first generates pseudo question-answer pairs in the target domain, and then uses the generated data as augmentation to fine-tune a pre-trained MRC model. However, the source-domain labeled data and target-domain pseudo data are directly combined without considering domain differences (see Figure \ref{fig:tsne}(a), where the two feature distributions in two domains are independently clustered). Directly transfering a model from one domain to another could be counter-effective, or even hurt the performance of the pre-trained model due to domain variance.

To achieve effective domain transfer, we need to learn features that are discriminative for the MRC task in the source domain, while simultaneously indiscriminating with respect to the shift between source and target domains.   
Motivated by this, we propose \emph{Adversarial Domain Adaptation for MRC} (AdaMRC), a new approach that utilizes adversarial learning to learn domain-invariant transferable representations for better MRC model adaptation across domains (see Figure \ref{fig:tsne}(b), where the two feature distributions learned by AdaMRC are indistinguishable through adversarial learning). 

Specifically, our proposed method first generates synthetic question-answer pairs given passages in the target domain. 
Different from~\citet{golub2017two}, which only used pseudo question-answer pairs to fine-tune pre-trained MRC models, our AdaMRC model uses the passage and the generated pseudo-questions in the target domain, in addition to the human-annotated passage-question pairs in the source domain, to train an additional \textit{domain classifier} as a discriminator. The passage-question encoder and the domain classifier are jointly trained via adversarial learning. In this way, the encoder is enforced to learn domain-invariant representations, which are beneficial for transferring knowledge learned from one domain to another. Based on this, an answer decoder is then used to decode domain-invariant representation into an answer span. 

The proposed approach is validated on a set of popular benchmarks, including SQuAD~\cite{rajpurkar2016squad}, NewsQA~\cite{trischler2016newsqa}, and MS MARCO~\cite{nguyen2016ms}, using state-of-the-art MRC models including SAN~\cite{P18-1157} and BiDAF~\cite{seo2016bidirectional}. 
Since pre-trained large-scale language models, such as ELMo~\cite{peters2018deep} and BERT~\cite{devlin2018bert}, have shown strong performance to learn representations that are generalizable to various tasks, in this work,  
to further demonstrate the versatility of the proposed model, we perform additional experiments to demonstrate that AdaMRC can also be combined with ELMo and BERT to further boost the performance.


The main contributions of this paper are summarized as follows: ($i$) We propose AdaMRC, an adversarial domain adaptation framework that is specifically designed for MRC. ($ii$) We perform comprehensive evaluations on several benchmarks, demonstrating that the proposed method is generalizable to different MRC models and diverse datasets. ($iii$) We demonstrate that AdaMRC is also compatible with ELMo and BERT. ($iv$) We further extend the proposed framework to semi-supervised learning, showing that AdaMRC can also be applied to boost the performance of a pre-trained MRC model when a small amount of labeled data is available in the target domain.

\section{Related Work}
\paragraph{Machine Reading Comprehension}
The MRC task has recently attracted a lot of attention in the community. An MRC system is required to answer a question by extracting a text snippet within a given passage as the answer. A large number of deep learning models have been proposed to tackle this task~\cite{seo2016bidirectional, xiong2016dynamic, shen2017reasonet, P18-1157, yu2018qanet}. However, the success of these methods largely relies on large-scale human-annotated datasets (such as SQuAD~\cite{rajpurkar2016squad}, NewsQA~\cite{trischler2016newsqa} and MS MARCO~\cite{nguyen2016ms}).

\begin{figure*}[t!]
\centering
\begin{tabular}{c}
\includegraphics[width=1.0\linewidth]{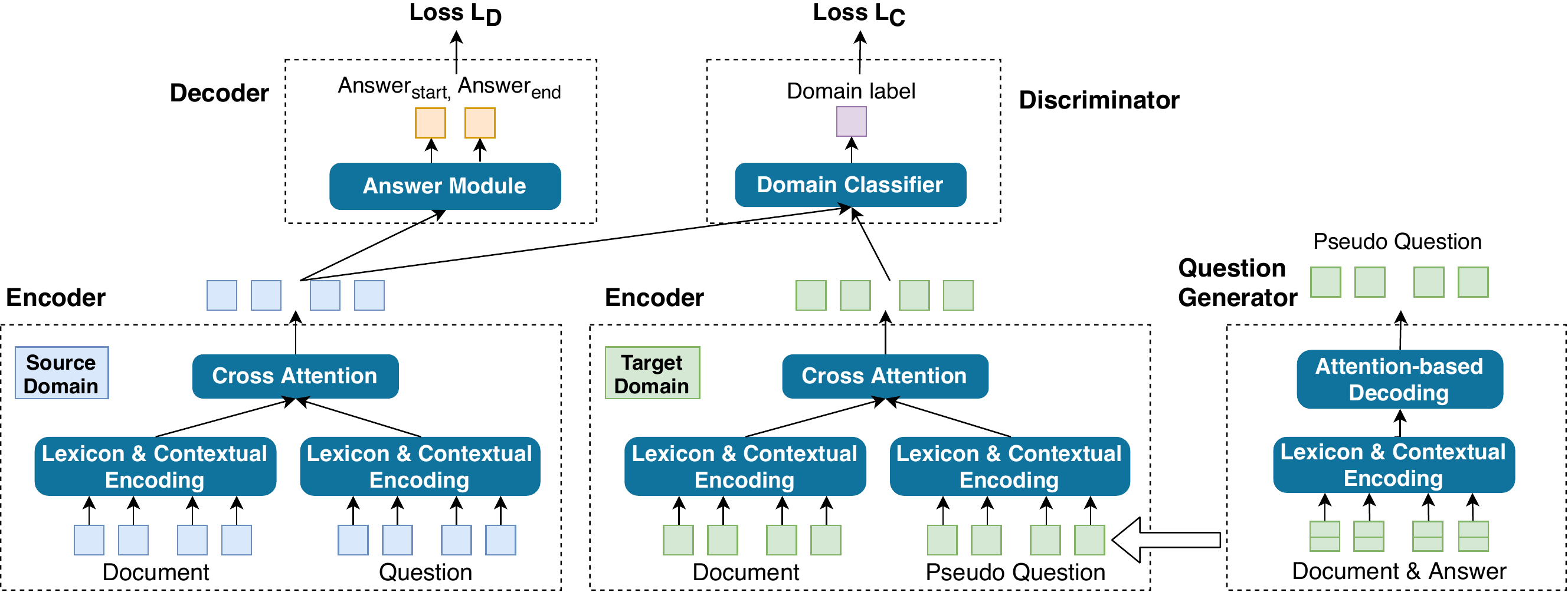} \\
\vspace{-6mm}
\end{tabular}
\caption{Illustration of the proposed AdaMRC model for unsupervised domain adaptation of MRC.}
\label{fig:model}
\end{figure*}

Different from previous work that focused on improving the state of the art on particular MRC datasets, we study the MRC task from a different angle, and aim at addressing a critical yet challenging problem: how to transfer an MRC model learned from a high-resource domain to other low-resource domains in an unsupervised manner. 

Although important for the MRC task, where annotated data are limited in real-life applications, this problem has not yet been well investigated.
There were some relevant studies along this line. For example, \citet{chung2017supervised} adapted a pre-trained model to TOEFL and MCTest dataset, and \citet{wiese2017neural} applied transfer learning to the biomedical domain. However, both studies assumed that annotated data in the target domain (either questions or question-answer pairs) are available. 

To the best of our knowledge, SynNet~\cite{golub2017two} is the only work that also studied domain adaptation for MRC. Compared with SynNet, the key difference in our model is adversarial learning, which enables domain-invariant representation learning for better model adaptation to low-resource domains. Our approach is also related to multi-task learning~\cite{xu2018multi,caruana1997multitask,liu2015representation,liu2019mt-dnn} and semi-supervised learning~\cite{yang2017semi} for MRC.
In this work, we focus on purely unsupervised domain adaptation. 

\paragraph{Domain Adaptation}
Domain adaptation aims to make a machine learning model generalizable to other domains, especially without any annotated data in the target domain (or with only limited data) \cite{ganin2015unsupervised}. One line of research on domain adaptation focuses on transiting the feature distribution from the source domain to the target domain \cite{gong2012geodesic, long2015learning}. Another school of research focuses on learning domain-invariant representations \cite{glorot2011domain} (e.g., via adversarial learning \cite{ganin2016domain, Tzeng_2017}).

Domain adaptation has been successfully applied to many tasks, such as image classification~\cite{Tzeng_2017}, speech recognition~\cite{doulaty2015data}, sentiment classification~\cite{ganin2016domain,li2017end}, machine translation~\cite{Johnson2017GooglesMN, zoph2016transfer},  relation extraction~\cite{fu2017domain}, and paraphrase identification~\cite{shah2018adversarial}. 
Compared to these areas, the application to MRC presents additional challenges, since besides missing labeled data (\emph{i.e.}, answer spans), the questions in the target domain are also unavailable. To our best knowledge, we are the first to investigate the usage of adversarial domain adaptation for the MRC task. 

There are many prevailing unsupervised techniques for domain adaptation. Our proposed approach is inspired by the seminal work of~\citet{ganin2016domain} to validate its potential of solving domain adaptation problem on a new task, without any supervision for the target domain. There are also other more advanced methods, such as MMD-based adaptation~\cite{long2017deep}, residual transfer network~\cite{long2016unsupervised}, and maximum classifier discrepancy~\cite{saito2018maximum} that can be explored for future work. 

\section{Problem Definition}
The problem of unsupervised domain adaptation for MRC is defined as follows. First, let $S=\{p^{s}, q^s, a^s\}$ denote a labeled MRC dataset from the source domain $s$, where $p^s, q^s$ and $a^s$ represent the passage, the question, and the answer of a sample, respectively. An MRC model $M^s$, taking as input the passage $p^s = (p_1, p_2, ... , p_T)$ of length $T$ and the question $q^s = (q_1, q_2, ... , q_{T'})$ of length $T'$, is trained to predict the correct answer span $a^s = (a_{start}^s, a_{end}^s)$, where $a_{start}^s, a_{end}^s$ represent the starting and ending indexes of the answer in the passage $p^s$. 

We assume that only unlabeled passages are available in the target domain $t$, \emph{i.e.}, $T = \{p^t\}$, where $p^t$ represents a passage. This is a reasonable assumption as it is easy to collect a large amount of unlabeled passages in a new domain. Given datasets $S$ and $T$, the goal of unsupervised domain adaptation is defined as learning an MRC model $M^t$ based on $S$ and $T$ to answer questions in the target domain $t$.  




\section{AdaMRC}
%

%
As illustrated in Figure \ref{fig:model}, AdaMRC consists of three main components: ($i$) \emph{Question Generator} (Sec.~\ref{sec:question_generation}), where pseudo question-answer pairs are generated given unlabeled passages in the target domain; ($ii$) \emph{MRC Module} (Sec.~\ref{sec:mrc}), 
where given an input document and a question, an answer span is extracted from the document; ($iii$) \emph{Domain Classifier} (Sec.~\ref{sec:domain_classifier}), where a domain label is predicted to distinguish a feature vector from either the source domain or the target domain. 

Specifically, the MRC module is composed of an encoder and a decoder. The encoder with parameter $\theta_e$ embeds the input passage and the question into a feature vector. The decoder with parameter $\theta_d$ takes the feature vector as input to predict the answer span. The domain classifier with parameter $\theta_c$ takes the same feature vector as input to classify the domain label. 
All the parameters $(\theta_e, \theta_d, \theta_c)$ are jointly optimized, with the objective of training the encoder to correctly predict the answer span, but also simultaneously fool the domain classifier. In other words, the encoder learns to map text input into a feature space that is invariant to the switch of domains. The following sub-sections describe each module, with training details provided in Sec.~\ref{sec:training}.






\subsection{Question Generation} \label{sec:question_generation}

First, we use an NER system to extract possible answer spans $a^t$ from the passages $p^t$ in the target domain, under the assumption that any named entity could be the potential answer of certain questions. Similar answer extraction strategy has been applied in \citet{yang2017semi} in a semi-supervised-learning setting, while \citet{golub2017two} proposed to train an answer synthesis network to predict possible answers spans. We tried both methods, and empirically observed that a simple NER system provides more robust results, which is used in our experiments.  


Now, we describe how the question generation (QG) model is trained. 
Given the passage $p^s = (p_1, p_2, ... , p_T)$ and answer $a^s = (a_{start}, a_{end})$ from the source domain, the QG model with parameter $\theta_{QG}$ learns the conditional probability of generating a question $q^s = (q_1, q_2, ... , q_{T'})$, \emph{i.e.}, $P(q^s|p^s,a^s)$. We implement the QG model as a sequence-to-sequence model with attention mechanism~\cite{bahdanau2014neural}, and also apply the copy mechanism proposed in~\citet{gu2016incorporating, gulcehre2016pointing} to handle rare/unknown words. 

Specifically, 
the QG model consists of a lexicon encoding layer, a BiLSTM contextual encoding layer, and an LSTM decoder. 
For lexicon encoding, each word token $p_i$ of a passage is mapped into a concatenation of GloVe vectors~\cite{pennington2014glove}, part-of-speech (POS) tagging embedding, and named-entity-recognition (NER) embedding. 
We further insert answer information by appending an additional zero/one feature (similar to \citet{yang2017semi}) to model the appearance of answer tokens in the passage. The output of the lexicon encoding layer is appended with CoVe vectors~\cite{mccann2017learned}, and then passed to the Bidirectional LSTM contextual encoding layer, producing a sequence of hidden states. 
The decoder is another LSTM with attention and copy mechanism over the encoder hidden states. At each time step, the generation probability of a question token $q_t$ is defined as:
\begin{align}
P(q_t) = g_t P^{v}(q_t) + (1-g_t)P^{copy}(q_t)\,,
\end{align}
where $g_t$ is the probability of generating a token from the vocabulary, while $(1-g_t)$ is the probability of copying a token from the passage. $P^{v}(q_t)$ and $P^{copy}(q_t)$ are defined as softmax functions over the words in the vocabulary and over the words in the passage, respectively. $g_t$, $P^v(q_t)$ and $P^{copy}(q_t)$ are functions of the current decoder hidden state.

\subsection{MRC Module}
\label{sec:mrc}
\paragraph{Encoder} The encoder in the MRC module contains lexicon encoding and contextual encoding, similar to the encoder used in the question generation module. It also includes a cross-attention layer for fusion. Specifically, the output of the lexicon encoder is appended with the CoVe vector and passed to the contextual encoding layer, which is a 2-layer Bidirectional LSTM that produces hidden states of the passage $H^p\in \R^{T\times 2m}$ and the question $H^q\in \R^{T'\times 2m}$, where $m$ is the hidden size of the BiLSTM. We then use cross attention to fuse $H^p$ and $H^q$, and construct a working memory of passage $M^p \in \R^{T\times 2m}$ (see \citet{P18-1157} for more details). The question memory $M^q \in \R^{2m}$ is constructed by applying self-attention on $H^q$. 

\paragraph{Decoder} The decoder, or answer module, predicts an answer span $a = (a_{start}, a_{end})$ given a passage $p$ and a question $q$, by modeling the conditional probability $P(a|p,q)$. The initial state $s_0$ is set as $M^q$. Through $T$ steps, a GRU~\cite{cho2014properties} is used to generate a sequence of state vectors $s_t=\text{GRU}(s_{t-1}, x_t)$, where $x_t$ is computed via attention between $M^p$ and $s_{t-1}$. Two softmax layers are used to compute the distribution of the start and the end of the answer span at each step given $s_t$, and the final prediction is the average prediction of all steps. Stochastic prediction dropout \cite{P18-1157} is applied during training.

Note that we use SAN as an example MRC model in the proposed framework. However, our approach is compatible with any existing MRC models. In experiments, in order to demonstrate the versatility of the proposed model, we also conduct experiments with BiDAF~\cite{seo2016bidirectional}.

\subsection{Domain Classifier} \label{sec:domain_classifier}

The domain classifier takes the output of the encoder as input, including the aforementioned passage representation $M^p \in \R^{T\times 2m}$ and the self-attended question representation $M^q \in \R^{2m}$ from different domains, and predicts the domain label $d$ by modeling the conditional probability $P(d|p, q)$. 
A self-attention layer is also applied to $M^p$ to reduce its size to $M^{p'} \in \R^{2m}$.  We then concatenate it with $M^q$, 
followed by a two-layer Multi-Layer Perceptron (MLP), $f(W[M^{p'}; M^q])$, and use a sigmoid function to predict the domain label.

\begin{algorithm}[t]
\caption{AdaMRC training procedure.} \label{Alg}
\begin{algorithmic}[1]
\State \textbf{Input:} source domain labeled data $S=\{p^s, q^s, a^s\} $, target domain unlabeled data $T=\{p^t\}$
\State Train the MRC model $\theta^s = (\theta_e^s, \theta_d^s)$ on source domain $S$; 
\State Train the QG model $\theta_{QG}$ on source domain $S$;
\State Generate $T_{gen}=\{p^t, q^t, a^t\} $ using the QG model;
\State Initialize $\theta = (\theta_e, \theta_d, \theta_c)$ with $\theta^s$;
\For {epoch $\leftarrow 1$ to $\#\text{epochs}$ }
\State Optimize $\theta$ on $S \cup T_{gen}$. Each minibatch is composed with $k_s$ samples from $S$ and $k_t$ samples from $T_{gen}$;
\EndFor
\State \textbf{Output:} Model with the best performance on the target development set $\theta^*$.
\end{algorithmic}
\end{algorithm}

\subsection{Training} \label{sec:training}
Algorithm \ref{Alg} illustrates the training procedure of our proposed framework. We first train the question generation model $\theta_{QG}$ on the source domain dataset $S$ by maximizing the likelihood of generating question $q^s$ given passage $p^s$ and answer $a^s$. Given the unlabeled dataset in the target domain, we extract candidate answers $a^t$ on $p^t$ and use $\theta_{QG}$ to generate pseudo questions $q^t$, and then compose a pseudo labeled dataset $T_{gen}= \{p^t, q^t, a^t\}$.

We initialize the MRC model $\theta$ for the target domain with the pre-trained MRC model $\theta^s$ from the source domain, and then fine-tune the model using both the source domain dataset $S$ and the target domain dataset $T_{gen}$. The goal of the decoder $\theta_d$ is to predict $P(a|p, q)$. The objective function is denoted as: 
\begin{align}
L_D(\theta_e, \theta_d) = \frac{1}{|S|}\textstyle{\sum_{i= 1}^{|S|}} \log P(a^{(i)}|p^{(i)}, q^{(i)})\,, 
\end{align}
where the superscript $(i)$ indicates the $i$-th sample.
It is worthwhile to emphasize that unlike \citet{golub2017two}, we only use source domain data to update the decoder, without using pseudo target domain data. This is because the synthetic question-answer pairs could be noisy, and directly using such data for decoder training may lead to degraded performance of the answer module, as observed both in \citet{sachan2018self} and in our experiments.

The synthetic target domain data and source domain data are both used to update the encoder $\theta_e$ and the domain classifier $\theta_c$. The classifier predicts a domain label $d$ given the feature representation from the encoder. The objective function is: %
\begin{align}
L_C(\theta_e, \theta_c) = \frac{1}{N}\textstyle{\sum_{i=1}^{N}} \log P(d^{(i)}|p^{(i)}, q^{(i)})\,, 
\end{align}
where $N=|S|+|T_{gen}|$.
In order to learn \textit{domain-invariant} representations from the encoder, we update $\theta_e$ to \textit{maximize} the loss while updating $\theta_c$ to minimize the loss in an adversarial fashion. The overall objective function is defined as: 
\begin{align}
    L(\theta_e, \theta_d, \theta_c) = L_D(\theta_e, \theta_d) - \lambda L_C(\theta_e, \theta_c)\,,
\end{align}
where $\lambda$ is a trade-off parameter that balances the two terms. 

To optimize our model,
instead of alternately updating the adversaries like in GAN~\cite{goodfellow2014generative}, 
we use the gradient-reversal layer~\cite{ganin2015unsupervised} to jointly optimize all the components, as suggested in \citet{chen2016adversarial}.  



\section{Experiments}
\label{sec:exp}

\subsection{Experimental Setting}
\label{subsec:setup}
\begin{table}[t!]
 \small
	\begin{center}
		\begin{tabular}{@{\hskip1pt}l@{\hskip1pt}|@{\hskip1pt}l@{\hskip1pt}|l@{\hskip1pt}|l@{\hskip1pt}|@{\hskip1pt}c@{\hskip1pt}}
			\hline \bf Dataset & Domain & Train & Dev & Test\\ \hline 
			SQuAD (v1.1) & Wiki  & 87,600& 10,570 & $-$\\ \hline
			NewsQA&  News & 92,549 & 5,166 & 5,165\\ \hline
			MS MARCO (v1) & Web&  82,430 &  10,047 & 9,650\\
			\hline
		\end{tabular}
	\end{center}
	\vspace{-3mm}
	\caption{Statistics of the datasets.}
	\vspace{-4mm}
	\label{tab:datasets}
\end{table}

\paragraph{Datasets} We validate our proposed method on three benchmarks: SQuAD~\cite{rajpurkar2016squad}, NewsQA~\cite{trischler2016newsqa}, and MS MARCO~\cite{nguyen2016ms}. The statistics of the datasets are provided in Table~\ref{tab:datasets}. Note that these datasets are all from different domains: SQuAD is from \textit{Wikipedia}; NewsQA is from \textit{CNN news}; and MS MARCO is from \textit{web search log}.


\paragraph{Evaluation metrics} For SQuAD and NewsQA, we report results on two evaluation metrics: Exact Match (\textbf{EM}), which measures the percentage of span predictions that match any of the ground truth
answers exactly; and Macro-averaged \textbf{F1} score,
which measures the average overlap between the
prediction and the ground-truth answer. For MS MARCO, since the answer is free-formed, we use BLEU and ROUGE-L scores for evaluation.  

\paragraph{Implementation details\footnote{Code will be released for easy access.}} We use spaCy\footnote{\url{https://spacy.io/}} to generate POS and NER taggings, which are used in answer extraction and the lexicon encoding layer of the QG and MRC models. The QG model is fixed after trained on source-domain labeled data. The hidden size of the LSTM in the QG model is set to 125. Parameters of the SAN model  follow \citet{P18-1157}. The hidden size of the MLP layer in the domain classifier is set to 125. Both the QG and the MRC model are optimized via Adamax \cite{kingma2014adam} with mini-batch size set to 32. The learning rate is set to 0.002 and is halved every 10 epochs. To avoid overfitting, we set the dropout rate to 0.3. For each mini-batch, data are sampled from both domains, with $k_s$ samples from the source domain and $k_t$ samples from the target domain. We set $k_s:k_t = 2:1$ as default in our experiments. For the trade-off parameter $\lambda$, we gradually change it from 0 to 1, following the schedule suggested in \citet{ganin2015unsupervised}.

\begin{table}[t]
    \small
    \centering
    \begin{tabular}{ll}
    \hline
        Method & EM/F1\\
        \hline
        \multicolumn{2}{c}{SQuAD $\rightarrow$ NewsQA}\\
        \hline
        SAN & 36.68/52.79 \\
        SynNet + SAN &35.19/49.61 \\
        \MN &\bf{38.46/54.20} \\
        \MN with GT questions &39.37,54.63 \\
        \hline
        \multicolumn{2}{c}{NewsQA $\rightarrow$ SQuAD}\\
        \hline
        SAN & 56.83/68.62\\
        SynNet + SAN &50.34/62.42\\
        \MN &\bf{58.20/69.75}\\
        \MN  with GT questions &58.82/70.14\\
        \hline
        \multicolumn{2}{c}{\small SQuAD $\rightarrow$ MS MARCO  (\small BLEU-1/ROUGE-L)}\\
        \hline
        SAN & 13.06/25.80  \\
        SynNet + SAN & 12.52/25.47 \\
        \MN &\bf{14.09/26.09} \\
        \MN  with GT questions & 15.59/26.40  \\
        \hline
        \multicolumn{2}{c}{MS MARCO $\rightarrow$ SQuAD}\\
        \hline
        SAN & 27.06/40.07 \\
        SynNet + SAN & 23.67/36.79 \\
        \MN &\bf{27.92/40.69}\\
        \MN with GT questions & 27.79/41.47 \\
        \hline        
    \end{tabular}
    \vspace{-2mm}
    \caption{Performance of \MN compared with baseline models on three datasets, using SAN as the MRC model.}\label{tab:san}
    \vspace{-3mm}
\end{table}

\subsection{Experimental Results}
\label{subsec:result}
We implement the following baselines and models for comparison. 
\begin{enumerate}
    \vspace{-2mm}
    \item \textbf{SAN}: we directly apply the pre-trained SAN model from the source domain to answer questions in the target domain.
    \vspace{-2mm}
    \item \textbf{SynNet+SAN}: we use SynNet\footnote{The officially released code is used in our experiments: https://github.com/davidgolub/QuestionGeneration.} \cite{golub2017two} to generate pseudo target-domain data, and then fine-tune the pre-trained SAN model.
    \vspace{-2mm}
    \item \textbf{AdaMRC}: as illustrated in Algorithm \ref{Alg}.
    \vspace{-2mm}
    \item \textbf{AdaMRC with GT questions}: the same as AdaMRC, except that the ground-truth questions in the target domain are used for training. This serves as an upper-bound of the proposed model.
\end{enumerate}

Table~\ref{tab:san} summarizes the experimental results. We observe that the proposed method consistently outperforms SAN and the SynNet+SAN model on all datasets. For example, in the SQuAD$\rightarrow$NewsQA setting, where the source-domain dataset is SQuAD and the target-domain dataset is NewsQA, AdaMRC achieves 38.46\% and 54.20\% in terms of EM and F1 scores, outperforming the pre-trained SAN by 1.78\% (EM) and 1.41\% (F1), respectively, as well as surpassing SynNet by 3.27\% (EM) and 4.59\% (F1), respectively.
Similar improvements are also observed in NewsQA$\rightarrow$SQuAD, SQuAD$\rightarrow$MS MARCO and MS MARCO$\rightarrow$SQuAD settings, which demonstrates the effectiveness of the proposed model. 

Interestingly, we find that the improvement on adaptation between  SQuAD and NewsQA is greater than that between SQuAD and MS MARCO. 
Our assumption is that it is because SQuAD and NewsQA datasets are more similar than SQuAD and MS MARCO, in terms of question style. For example, questions in MS MARCO are real web search queries, which are short and may have typos or abbreviations; while questions in SQuAD and NewsQA are more formal and well written. Furthermore, the ground-truth answers in MS MARCO are human-synthesized and usually much longer (16.4 tokens in average) than those in the other datasets, while our answer extraction process focuses on named entities (which are much shorter). 
We argue that extracting named entities as possible answers is still reasonable for most of the reading comprehension tasks such as SQuAD and NewsQA. The problem of synthesizing answers across different domains will be investigated in future work. 



\paragraph{SynNet vs. pre-trained SAN baseline}

One observation is that SynNet performs worse than the pre-trained SAN baseline. We hypothesize that this is because the generated question-answer pairs are often noisy and inaccurate, and directly fine-tuning the answer module with synthetic data may hurt the performance, which is also observed in \citet{sachan2018self}, especially when a well-performed MRC model is used as the baseline. Note that we do observe improvements from SynNet+BiDAF over the pre-trained BiDAF model, which will be discussed in Sec. \ref{sec:bidaf}. 

\paragraph{Comparing with upper-bound}

The ``AdaMRC with GT questions'' model (in Section \ref{subsec:result}) serves as the upper-bound of our proposed approach, where ground-truth questions are used instead of synthesized questions. By using ground-truth questions, performance is further boosted by around 1\%. This suggests that our question generation model is effective as the margin is relatively small, yet it could be further improved. We plan to study if recent question generation methods \cite{du2017learning, duan2017question,sun2018answer,benmalek2019keeping} could further help to close the performance gap in future work.


\section{Analysis}
\subsection{Visualization} 
To demonstrate the effectiveness of adversarial domain adaptation, we visualize the encoded representation via t-SNE~\cite{maaten2008visualizing} in Figure \ref{fig:tsne}. 
We observe that with AdaMRC, the two distributions of encoded feature representations are indistinguishable.
Without AdaMRC, the two distributions are independently clustered by domain. We further use KL divergence for measuring distributional differences. The KL divergence of data samples between source and target domains, with and without domain adaptation, are 0.278,
0.433, respectively (smaller is better).


\subsection{Robustness of \MN} \label{sec:bidaf}


\begin{figure}[t]
    \centering
\subfigure[From SQuAD to NewsQA.]{
\adjustbox{trim={.01\width} {.01\height} {.0\width} {.01\height},clip}{\includegraphics[scale=0.25]{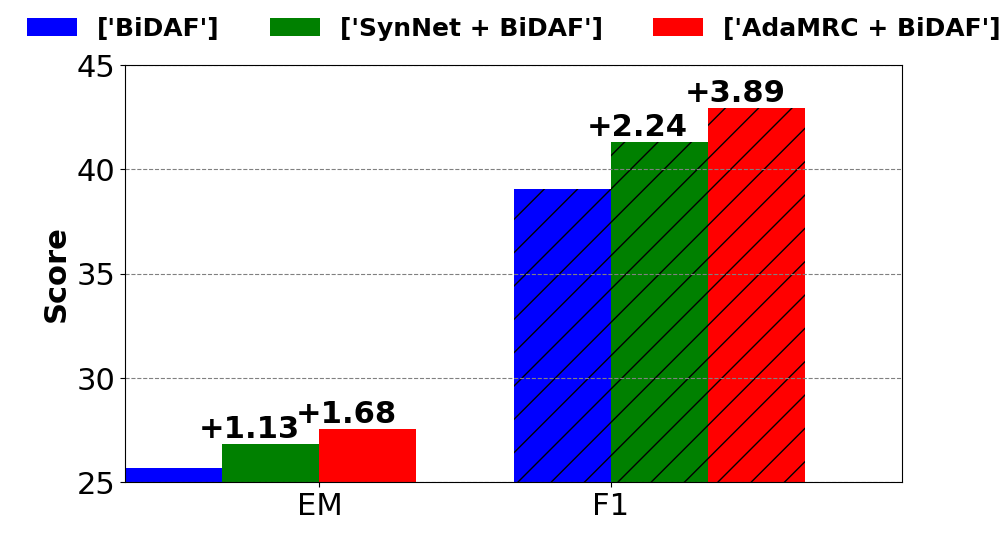}}}
\vspace{-3mm}
\subfigure[From NewsQA to SQuAD.]{
\adjustbox{trim={.01\width} {.01\height} {.01\width} {.01\height},clip}{\includegraphics[scale=0.25]{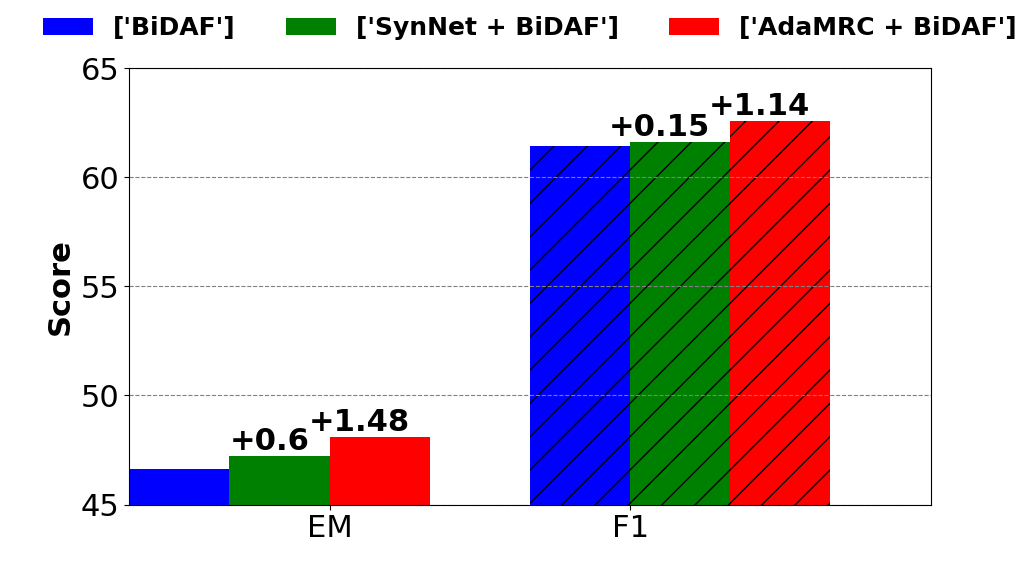}}}
\caption{\label{fig:comp_bidaf} Performance of our proposed method compared with baselines, using BiDAF as the MRC model.
} 
\vspace{-3mm}
\end{figure} 
\begin{table}[t]
    \small
    \centering
    \begin{tabular}{ll}
    \hline
        Method & EM/F1 \\
        \hline
        SAN & 32.35/42.62 \\
        AdaMRC + SAN &\bf{33.61/44.16} \\
        \hline
        BiDAF & 27.85/36.82 \\
        AdaMRC + BiDAF &\bf{29.12/38.84 } \\  
        \hline
    \end{tabular}
    
    \caption{Performance on DuoRC, adapting from SelfRC (Wikipedia) to ParaphraseRC (IMDB).}\label{tab:duorc}
    \vspace{-3mm}
\end{table}  


\paragraph{Results on BiDAF}  To verify that our proposed framework is compatible to existing MRC models, we also apply our framework to the BiDAF model, which has different encoder and decoder structures compared to SAN. We follow the model architecture and parameter settings in \citet{seo2016bidirectional}.  
As shown in Figure \ref{fig:comp_bidaf}, the proposed AdaMRC model clearly outperforms both SynNet+BiDAF and pre-trained BiDAF model.
We also observe that the improvement of AdaMRC over BiDAF is more significant than SAN. Our hypothesis is that since BiDAF is a weaker baseline than SAN, a higher performance improvement can be observed when the domain adaptation approach is applied to enhance the model. This experiment confirms that our proposed approach is robust and can generalize to different MRC models.

   

\paragraph{Results on DuoRC} 
We further test our model on the newly-released DuoRC dataset \cite{saha2018duorc}. This dataset contains two subsets: movie descriptions collected from \textit{Wikipedia} (SelfRC) and from \textit{IMDB} (ParaphraseRC). Although the two subsets are describing the same movies, the documents from Wikipedia are usually shorter (580 words in average), while the documents from IMDB are longer and more descriptive (926 words in average). We consider them as two different domains and perform domain adaptation from Wikipedia to IMDB. This experiment broadens our definition of domain.

In the DuoRC dataset, the same questions are asked on both Wikipedia and IMDB documents. Thus, question synthesis is not needed, and comparison with SynNet is not feasible. Note that the answers of the same question could be different in the two subsets (only 40.7\% of the questions have the same answers in both domains). We pre-process the dataset and test the answer-span extraction task following \citet{saha2018duorc}. Results are reported in Table \ref{tab:duorc}. AdaMRC improves the performance over both SAN (1.26\%, 1.54\% in EM and F1) and BiDAF (1.27\%, 2.02\% in EM and F1). 
This experiment validates that our method can be applied to different styles of domain adaptation tasks as well. 

\begin{table}[t]
    \small
    \centering
    \begin{tabular}{ll}
    \hline
        Method & EM/F1 \\
        \hline
        SAN & 36.68/52.79 \\
        AdaMRC + SAN &\bf{38.46/54.20} \\ 
        \hline
        SAN + ELMo& 39.61/55.18 \\
        AdaMRC + SAN + ELMo &\bf{40.96/56.25} \\  
        \hline
        $\text{BERT}_{\text{BASE}}$ & 42.00/58.71\\
        AdaMRC + $\text{BERT}_{\text{BASE}}$ & \bf{42.59/59.25}\\
        \hline
    \end{tabular}
   
    \caption{Results of using ELMo and BERT. Setting: adaptation from SQuAD to NewsQA.}\label{tab:elmo}
     \vspace{-2mm}
\end{table} 

\subsection{AdaMRC with Pre-trained Language Models}
To verify that our approach is compatible with large-scale pre-trained language models, we evaluate our model with ELMo~\cite{peters2018deep} and BERT~\cite{devlin2018bert}. To apply ELMo to SAN, we use the model provided by AllenNLP\footnote{\url{https://allennlp.org/}}, and append a 1024-dim ELMo vector to the contextual encoding layer, with dropout rate set to 0.5. For BERT, we experiment with the pre-trained $\text{BERT}_{\text{BASE}}$ uncased model\footnote{\url{https://github.com/google-research/bert}} due to limited computational resources. We use the original design of finetuning BERT for the MRC task in \citet{devlin2018bert}, instead of combining BERT with SAN. Results are provided in Table \ref{tab:elmo}. We observe that using ELMo and BERT improves both AdaMRC and the baseline model. However, the improvement over ELMo and BERT is relatively smaller than SAN. We believe this is because pre-trained language model provides additional domain-invariant information learned from external data, and therefore limits the improvement of domain-invariant feature learning in our model. However, it is worth noting that combining AdaMRC with BERT achieves the best performance, which validates that AdaMRC is compatible with data augmentation from external sources.

\begin{table}[t]
    \small
    \centering
    \begin{tabular}{ccc}
    \hline
        Ratio &  SAN & AdaMRC + SAN\\
        \hline
        0\%& 36.68/52.79 & \bf{38.46/54.20}\\
        5\% & 47.61/62.69& \bf{48.50/63.17}\\
        10\% & 48.66/63.32& \bf{49.64/63.94}\\
        20\% & 50.75/64.80& \bf{51.14/65.38}\\
        50\% &  53.24/67.07& \bf{53.34/67.30}\\
        100\% & \bf{56.48/69.14}& 56.29/68.97 \\
        \hline
    \end{tabular}
    \caption{Semi-supervised domain adaptation experiment with varied labeling ratio on the target-domain dataset. Setting: adaptation from SQuAD to NewsQA.}\label{tab:semi}
    \vspace{-2mm}
\end{table} 

\begin{table*}[t!]
    \small
    \centering
    \vspace{-2mm}
    \begin{tabulary}{\linewidth}{L}
    \hline \hline
    \noalign{\vskip 1mm} 
    \textit{Refugee camps in eastern Chad house about 300,000 people who fled violence in the Darfur region of Sudan . The U.N. High Commissioner for Refugees said on Monday that more than \textbf{\color{blue}12,000} people have fled militia attacks over the last few days from Sudan 's Darfur region to neighboring Chad}... \\
    \textbf{Answer: \color{blue}{12,000}} \\
    \textbf{GT Question}: How many have recently crossed to Chad? \\
    \textbf{Pseudo Question}: How many people fled the Refugee region to Sudan?\\   
    \noalign{\vskip 1mm} 
    \hline \hline
    \noalign{\vskip 1mm} 
    \textit{Sources say the classified materials were taken from the East Tennessee Technology Park . \textbf{\color{blue}Roy Lynn Oakley} , 67 , of Roane County , Tennessee , appeared in federal court in Knoxville on Thursday . Oakley was briefly detained for questioning in the case in January} ...\\
    \textbf{Answer: \color{blue}{Roy Lynn Oakley}} \\
    \textbf{GT Question}: Who is appearing in court ? \\
    \textbf{Pseudo Question}: What is the name of the classified employee in Tennessee on East Tennessee ?\\  
    \noalign{\vskip 1mm} 
    \hline \hline
    \noalign{\vskip 1mm} 
    \textit{The Kyrgyz order became effective on Friday when \textbf{\color{blue}President Kurmanbek Bakiyev} reportedly signed legislation that the parliament in Bishkek backed on Thursday , the Pentagon said . Pentagon spokesman Bryan Whitman said the Kyrgyz Foreign Ministry on Friday officially notified the U.S. Embassy in Bishkek that a 180-day withdrawal process is under way}...\\
    \textbf{Answer: \color{blue}President Kurmanbek Bakiyev} \\
    \textbf{GT Question}: Who is the President of Kyrgyzstan ? \\
    \textbf{Pseudo Question}: What spokesman signed legislation that the parliament was signed legislation in 2011 ?\\  
    \noalign{\vskip 1mm} 
    \hline \hline
    \noalign{\vskip 1mm} 
    \textit{A high court in northern India on Friday acquitted a wealthy businessman facing the death sentence for the killing of a teen in a case dubbed " the house of horrors . " Moninder Singh Pandher was sentenced to death by a lower court in February . The teen was \textbf{\color{blue}one of 19} victims -- children and young women -- in one of the most gruesome serial killings in India in recent years} ...\\
    \textbf{Answer: \color{blue}{one of 19}}\\
    \textbf{GT Question}:What was the amount of children murdered?\\
    \textbf{Pseudo Question}: How many victims were in India?\\  
    \noalign{\vskip 1mm} 
    \hline \hline
    \end{tabulary}
    \caption{Examples of generated questions given input paragraphs and answers, comparing with the ground-truth human-written questions.}\label{tab:questions}
    \vspace{-4mm}
\end{table*} 

\subsection{Semi-supervised Setting}
As an additional experiment, we also evaluate the proposed AdaMRC framework for semi-supervised domain adaptation. We randomly sample $k$ portion of labeled data from the target domain, and feed them to the MRC model. The ratio of labeled data ranges $k$ from 0\% to 100\%. Table \ref{tab:semi} shows that AdaMRC outperforms SAN. However, the gap is decreasing when the labeling ratio increases. When the ratio is 20\% or smaller, there is noticeable improvement. When the ratio is set to 50\%, the two methods result in similar performance. When the ratio is increased to 100\%, \emph{i.e.}, fully supervised learning, the performance of AdaMRC is slightly worse than SAN. This is possibly because in a supervised learning setting, the encoder is trained to preserve domain-specific feature information. 
The overall results suggest that our proposed AdaMRC is also effective in semi-supervised setting, when a small portion of target-domain data is provided. 

\subsection{Examples of Generated Questions}
The percentage of generated questions starting with ``what'', ``who'',
``when'' and ``where'' are 63.2\%, 12.8\%, 2.3\%, and 2.1\%, respectively.
We provide several examples of generated
questions in Table \ref{tab:questions}. 
We observe that the generated questions are longer than human-written questions. This is possibly due to the copy mechanism used in the question generation model, which enables directly copying words into the generated questions. On the one hand, the copy mechanism provides detailed background information for generating a question. However, if not copying correctly, the question could be syntactically incorrect. For instance, in the third example, ``\emph{signed legislation that the parliament}'' is copied from the passage. The copied phrase is indeed describing the answer ``\emph{President Kurmanbek Bakiyev}''; however, the question is syntactically incorrect and the question generator should copy ``\emph{the parliament backed on Thursday}'' instead.

There is generally good correspondence between the answer type and generated questions. For example, the question generator will produce ``\emph{What is the name of}'' if the answer is about a person, and  ask ``\emph{How many}'' if the answer is a number. We also observe that the generated questions may encounter semantic errors though syntactically fluent. For instance, in the first example, the passage suggests that people fled \emph{from} Sudan to Chad, while the generated question describes the wrong direction. However, overall we think that the current question generator provides reasonable synthesized questions, yet there is still large room to improve. The observation also confirms our analysis that the synthetic question-answer pairs could be noisy and inaccurate, thus could hurt the performance when fine-tuning the answer module with generated data.

\section{Conclusion}
%
In this paper, we propose a new framework, \emph{Adversarial Domain Adaptation for MRC} (AdaMRC), to transfer a pre-trained  MRC  model from a source domain to a target domain. We validate our proposed framework on several datasets and observe consistent improvement over baseline methods. We also verify the robustness of the proposed framework by applying it to different MRC models. Experiments also show that AdaMRC is compatible with pre-trained language model and semi-supervised learning setting. 
We believe our analysis provides insights that can help guide further research in this task.  

\clearpage
\bibliography{references}

\begin{thebibliography}{51}
\expandafter\ifx\csname natexlab\endcsname\relax\def\natexlab#1{#1}\fi

\bibitem[{Bahdanau et~al.(2015)Bahdanau, Cho, and Bengio}]{bahdanau2014neural}
Dzmitry Bahdanau, Kyunghyun Cho, and Yoshua Bengio. 2015.
\newblock Neural machine translation by jointly learning to align and
  translate.
\newblock In \emph{ICLR}.

\bibitem[{Bajaj et~al.(2016)Bajaj, Campos, Craswell, Deng, Gao, Liu, Majumder,
  McNamara, Mitra, Nguyen et~al.}]{nguyen2016ms}
Payal Bajaj, Daniel Campos, Nick Craswell, Li~Deng, Jianfeng Gao, Xiaodong Liu,
  Rangan Majumder, Andrew McNamara, Bhaskar Mitra, Tri Nguyen, et~al. 2016.
\newblock Ms marco: A human generated machine reading comprehension dataset.
\newblock \emph{arXiv preprint arXiv:1611.09268}.

\bibitem[{Benmalek et~al.(2019)Benmalek, Khabsa, Desu, Cardie, and
  Banko}]{benmalek2019keeping}
Ryan Benmalek, Madian Khabsa, Suma Desu, Claire Cardie, and Michele Banko.
  2019.
\newblock Keeping notes: Conditional natural language generation with a
  scratchpad encoder.
\newblock In \emph{ACL}.

\bibitem[{Caruana(1997)}]{caruana1997multitask}
Rich Caruana. 1997.
\newblock Multitask learning.
\newblock \emph{Machine learning}.

\bibitem[{Chen et~al.(2018)Chen, Sun, Athiwaratkun, Cardie, and
  Weinberger}]{chen2016adversarial}
Xilun Chen, Yu~Sun, Ben Athiwaratkun, Claire Cardie, and Kilian Weinberger.
  2018.
\newblock Adversarial deep averaging networks for cross-lingual sentiment
  classification.
\newblock \emph{TACL}.

\bibitem[{Cho et~al.(2014)Cho, Van~Merri{\"e}nboer, Bahdanau, and
  Bengio}]{cho2014properties}
Kyunghyun Cho, Bart Van~Merri{\"e}nboer, Dzmitry Bahdanau, and Yoshua Bengio.
  2014.
\newblock On the properties of neural machine translation: Encoder-decoder
  approaches.
\newblock \emph{arXiv preprint arXiv:1409.1259}.

\bibitem[{Chung et~al.(2018)Chung, Lee, and Glass}]{chung2017supervised}
Yu-An Chung, Hung-Yi Lee, and James Glass. 2018.
\newblock Supervised and unsupervised transfer learning for question answering.
\newblock In \emph{NAACL}.

\bibitem[{Devlin et~al.(2019)Devlin, Chang, Lee, and
  Toutanova}]{devlin2018bert}
Jacob Devlin, Ming-Wei Chang, Kenton Lee, and Kristina Toutanova. 2019.
\newblock Bert: Pre-training of deep bidirectional transformers for language
  understanding.
\newblock In \emph{NAACL}.

\bibitem[{Doulaty et~al.(2015)Doulaty, Saz, and Hain}]{doulaty2015data}
Mortaza Doulaty, Oscar Saz, and Thomas Hain. 2015.
\newblock Data-selective transfer learning for multi-domain speech recognition.
\newblock \emph{arXiv preprint arXiv:1509.02409}.

\bibitem[{Du et~al.(2017)Du, Shao, and Cardie}]{du2017learning}
Xinya Du, Junru Shao, and Claire Cardie. 2017.
\newblock Learning to ask: Neural question generation for reading
  comprehension.
\newblock In \emph{ACL}.

\bibitem[{Duan et~al.(2017)Duan, Tang, Chen, and Zhou}]{duan2017question}
Nan Duan, Duyu Tang, Peng Chen, and Ming Zhou. 2017.
\newblock Question generation for question answering.
\newblock In \emph{EMNLP}.

\bibitem[{Fu et~al.(2017)Fu, Nguyen, Min, and Grishman}]{fu2017domain}
Lisheng Fu, Thien~Huu Nguyen, Bonan Min, and Ralph Grishman. 2017.
\newblock Domain adaptation for relation extraction with domain adversarial
  neural network.
\newblock In \emph{IJCNLP}.

\bibitem[{Ganin and Lempitsky(2015)}]{ganin2015unsupervised}
Yaroslav Ganin and Victor Lempitsky. 2015.
\newblock Unsupervised domain adaptation by backpropagation.
\newblock In \emph{ICML}.

\bibitem[{Ganin et~al.(2016)Ganin, Ustinova, Ajakan, Germain, Larochelle,
  Laviolette, Marchand, and Lempitsky}]{ganin2016domain}
Yaroslav Ganin, Evgeniya Ustinova, Hana Ajakan, Pascal Germain, Hugo
  Larochelle, Fran{\c{c}}ois Laviolette, Mario Marchand, and Victor Lempitsky.
  2016.
\newblock Domain-adversarial training of neural networks.
\newblock \emph{JMLR}.

\bibitem[{Gao et~al.(2019)Gao, Galley, and Li}]{gaosurvey}
Jianfeng Gao, Michel Galley, and Lihong Li. 2019.
\newblock Neural approaches to conversational ai.
\newblock \emph{Foundations and Trends{\textregistered} in Information
  Retrieval}.

\bibitem[{Glorot et~al.(2011)Glorot, Bordes, and Bengio}]{glorot2011domain}
Xavier Glorot, Antoine Bordes, and Yoshua Bengio. 2011.
\newblock Domain adaptation for large-scale sentiment classification: A deep
  learning approach.
\newblock In \emph{ICML}.

\bibitem[{Golub et~al.(2017)Golub, Huang, He, and Deng}]{golub2017two}
David Golub, Po-Sen Huang, Xiaodong He, and Li~Deng. 2017.
\newblock Two-stage synthesis networks for transfer learning in machine
  comprehension.
\newblock In \emph{EMNLP}.

\bibitem[{Gong et~al.(2012)Gong, Shi, Sha, and Grauman}]{gong2012geodesic}
Boqing Gong, Yuan Shi, Fei Sha, and Kristen Grauman. 2012.
\newblock Geodesic flow kernel for unsupervised domain adaptation.
\newblock In \emph{CVPR}.

\bibitem[{Goodfellow et~al.(2014)Goodfellow, Pouget-Abadie, Mirza, Xu,
  Warde-Farley, Ozair, Courville, and Bengio}]{goodfellow2014generative}
Ian Goodfellow, Jean Pouget-Abadie, Mehdi Mirza, Bing Xu, David Warde-Farley,
  Sherjil Ozair, Aaron Courville, and Yoshua Bengio. 2014.
\newblock Generative adversarial nets.
\newblock In \emph{NIPS}.

\bibitem[{Gu et~al.(2016)Gu, Lu, Li, and Li}]{gu2016incorporating}
Jiatao Gu, Zhengdong Lu, Hang Li, and Victor~OK Li. 2016.
\newblock Incorporating copying mechanism in sequence-to-sequence learning.
\newblock In \emph{ACL}.

\bibitem[{Gulcehre et~al.(2016)Gulcehre, Ahn, Nallapati, Zhou, and
  Bengio}]{gulcehre2016pointing}
Caglar Gulcehre, Sungjin Ahn, Ramesh Nallapati, Bowen Zhou, and Yoshua Bengio.
  2016.
\newblock Pointing the unknown words.
\newblock In \emph{ACL}.

\bibitem[{Johnson et~al.(2017)Johnson, Schuster, Le, Krikun, Wu, Chen, Thorat,
  Vi{\'e}gas, Wattenberg, Corrado, Hughes, and Dean}]{Johnson2017GooglesMN}
Melvin Johnson, Mike Schuster, Quoc~V. Le, Maxim Krikun, Yonghui Wu, Zhifeng
  Chen, Nikhil Thorat, Fernanda~B. Vi{\'e}gas, Martin Wattenberg, Gregory~S.
  Corrado, Macduff Hughes, and Jeffrey Dean. 2017.
\newblock Google's multilingual neural machine translation system: Enabling
  zero-shot translation.
\newblock \emph{TACL}.

\bibitem[{Kingma and Ba(2015)}]{kingma2014adam}
Diederik~P Kingma and Jimmy Ba. 2015.
\newblock Adam: A method for stochastic optimization.
\newblock In \emph{ICLR}.

\bibitem[{Li et~al.(2017)Li, Zhang, Wei, Wu, and Yang}]{li2017end}
Zheng Li, Yun Zhang, Ying Wei, Yuxiang Wu, and Qiang Yang. 2017.
\newblock End-to-end adversarial memory network for cross-domain sentiment
  classification.
\newblock In \emph{IJCAI}.

\bibitem[{Liu et~al.(2015)Liu, Gao, He, Deng, Duh, and
  Wang}]{liu2015representation}
Xiaodong Liu, Jianfeng Gao, Xiaodong He, Li~Deng, Kevin Duh, and Ye-Yi Wang.
  2015.
\newblock Representation learning using multi-task deep neural networks for
  semantic classification and information retrieval.
\newblock In \emph{NAACL}.

\bibitem[{Liu et~al.(2019)Liu, He, Chen, and Gao}]{liu2019mt-dnn}
Xiaodong Liu, Pengcheng He, Weizhu Chen, and Jianfeng Gao. 2019.
\newblock Multi-task deep neural networks for natural language understanding.
\newblock In \emph{ACL}.

\bibitem[{Liu et~al.(2018)Liu, Shen, Duh, and Gao}]{P18-1157}
Xiaodong Liu, Yelong Shen, Kevin Duh, and Jianfeng Gao. 2018.
\newblock Stochastic answer networks for machine reading comprehension.
\newblock In \emph{ACL}.

\bibitem[{Long et~al.(2015)Long, Cao, Wang, and Jordan}]{long2015learning}
Mingsheng Long, Yue Cao, Jianmin Wang, and Michael~I. Jordan. 2015.
\newblock Learning transferable features with deep adaptation networks.
\newblock In \emph{ICML}.

\bibitem[{Long et~al.(2016)Long, Zhu, Wang, and Jordan}]{long2016unsupervised}
Mingsheng Long, Han Zhu, Jianmin Wang, and Michael~I Jordan. 2016.
\newblock Unsupervised domain adaptation with residual transfer networks.
\newblock In \emph{NIPS}.

\bibitem[{Long et~al.(2017)Long, Zhu, Wang, and Jordan}]{long2017deep}
Mingsheng Long, Han Zhu, Jianmin Wang, and Michael~I Jordan. 2017.
\newblock Deep transfer learning with joint adaptation networks.
\newblock In \emph{ICML}.

\bibitem[{Maaten and Hinton(2008)}]{maaten2008visualizing}
Laurens van~der Maaten and Geoffrey Hinton. 2008.
\newblock Visualizing data using t-sne.
\newblock \emph{JMLR}.

\bibitem[{McCann et~al.(2017)McCann, Bradbury, Xiong, and
  Socher}]{mccann2017learned}
Bryan McCann, James Bradbury, Caiming Xiong, and Richard Socher. 2017.
\newblock Learned in translation: Contextualized word vectors.
\newblock In \emph{NIPS}.

\bibitem[{Pennington et~al.(2014)Pennington, Socher, and
  Manning}]{pennington2014glove}
Jeffrey Pennington, Richard Socher, and Christopher Manning. 2014.
\newblock Glove: Global vectors for word representation.
\newblock In \emph{EMNLP}.

\bibitem[{Peters et~al.(2018)Peters, Neumann, Iyyer, Gardner, Clark, Lee, and
  Zettlemoyer}]{peters2018deep}
Matthew~E Peters, Mark Neumann, Mohit Iyyer, Matt Gardner, Christopher Clark,
  Kenton Lee, and Luke Zettlemoyer. 2018.
\newblock Deep contextualized word representations.
\newblock In \emph{NAACL}.

\bibitem[{Rajpurkar et~al.(2016)Rajpurkar, Zhang, Lopyrev, and
  Liang}]{rajpurkar2016squad}
Pranav Rajpurkar, Jian Zhang, Konstantin Lopyrev, and Percy Liang. 2016.
\newblock Squad: 100,000+ questions for machine comprehension of text.
\newblock In \emph{EMNLP}.

\bibitem[{Sachan and Xing(2018)}]{sachan2018self}
Mrinmaya Sachan and Eric Xing. 2018.
\newblock Self-training for jointly learning to ask and answer questions.
\newblock In \emph{NAACL}.

\bibitem[{Saha et~al.(2018)Saha, Aralikatte, Khapra, and
  Sankaranarayanan}]{saha2018duorc}
Amrita Saha, Rahul Aralikatte, Mitesh~M Khapra, and Karthik Sankaranarayanan.
  2018.
\newblock Duorc: Towards complex language understanding with paraphrased
  reading comprehension.
\newblock In \emph{ACL}.

\bibitem[{Saito et~al.(2018)Saito, Watanabe, Ushiku, and
  Harada}]{saito2018maximum}
Kuniaki Saito, Kohei Watanabe, Yoshitaka Ushiku, and Tatsuya Harada. 2018.
\newblock Maximum classifier discrepancy for unsupervised domain adaptation.
\newblock In \emph{CVPR}.

\bibitem[{Seo et~al.(2017)Seo, Kembhavi, Farhadi, and
  Hajishirzi}]{seo2016bidirectional}
Minjoon Seo, Aniruddha Kembhavi, Ali Farhadi, and Hannaneh Hajishirzi. 2017.
\newblock Bidirectional attention flow for machine comprehension.
\newblock In \emph{ICLR}.

\bibitem[{Shah et~al.(2018)Shah, Lei, Moschitti, Romeo, and
  Nakov}]{shah2018adversarial}
Darsh~J Shah, Tao Lei, Alessandro Moschitti, Salvatore Romeo, and Preslav
  Nakov. 2018.
\newblock Adversarial domain adaptation for duplicate question detection.
\newblock In \emph{EMNLP}.

\bibitem[{Shen et~al.(2017)Shen, Huang, Gao, and Chen}]{shen2017reasonet}
Yelong Shen, Po-Sen Huang, Jianfeng Gao, and Weizhu Chen. 2017.
\newblock Reasonet: Learning to stop reading in machine comprehension.
\newblock In \emph{KDD}.

\bibitem[{Sun et~al.(2018)Sun, Liu, Lyu, He, Ma, and Wang}]{sun2018answer}
Xingwu Sun, Jing Liu, Yajuan Lyu, Wei He, Yanjun Ma, and Shi Wang. 2018.
\newblock Answer-focused and position-aware neural question generation.
\newblock In \emph{EMNLP}.

\bibitem[{Trischler et~al.(2016)Trischler, Wang, Yuan, Harris, Sordoni,
  Bachman, and Suleman}]{trischler2016newsqa}
Adam Trischler, Tong Wang, Xingdi Yuan, Justin Harris, Alessandro Sordoni,
  Philip Bachman, and Kaheer Suleman. 2016.
\newblock Newsqa: A machine comprehension dataset.
\newblock \emph{arXiv preprint arXiv:1611.09830}.

\bibitem[{Tzeng et~al.(2017)Tzeng, Hoffman, Saenko, and Darrell}]{Tzeng_2017}
Eric Tzeng, Judy Hoffman, Kate Saenko, and Trevor Darrell. 2017.
\newblock Adversarial discriminative domain adaptation.
\newblock In \emph{CVPR}.

\bibitem[{Wiese et~al.(2017)Wiese, Weissenborn, and Neves}]{wiese2017neural}
Georg Wiese, Dirk Weissenborn, and Mariana Neves. 2017.
\newblock Neural question answering at bioasq 5b.
\newblock In \emph{BioNLP workshop}.

\bibitem[{Xiong et~al.(2017)Xiong, Zhong, and Socher}]{xiong2016dynamic}
Caiming Xiong, Victor Zhong, and Richard Socher. 2017.
\newblock Dynamic coattention networks for question answering.
\newblock In \emph{ICLR}.

\bibitem[{Xu et~al.(2019)Xu, Liu, Shen, Liu, and Gao}]{xu2018multi}
Yichong Xu, Xiaodong Liu, Yelong Shen, Jingjing Liu, and Jianfeng Gao. 2019.
\newblock Multi-task learning with sample re-weighting for machine reading
  comprehension.
\newblock In \emph{NAACL}.

\bibitem[{Yang et~al.(2017)Yang, Hu, Salakhutdinov, and Cohen}]{yang2017semi}
Zhilin Yang, Junjie Hu, Ruslan Salakhutdinov, and William Cohen. 2017.
\newblock Semi-supervised qa with generative domain-adaptive nets.
\newblock In \emph{ACL}.

\bibitem[{Yu et~al.(2018)Yu, Dohan, Luong, Zhao, Chen, Norouzi, and
  Le}]{yu2018qanet}
Adams~Wei Yu, David Dohan, Minh-Thang Luong, Rui Zhao, Kai Chen, Mohammad
  Norouzi, and Quoc~V Le. 2018.
\newblock Qanet: Combining local convolution with global self-attention for
  reading comprehension.
\newblock In \emph{ICLR}.

\bibitem[{Zhang et~al.(2018)Zhang, Liu, Liu, Gao, Duh, and
  Van~Durme}]{zhang2018record}
Sheng Zhang, Xiaodong Liu, Jingjing Liu, Jianfeng Gao, Kevin Duh, and Benjamin
  Van~Durme. 2018.
\newblock Record: Bridging the gap between human and machine commonsense
  reading comprehension.
\newblock \emph{arXiv preprint arXiv:1810.12885}.

\bibitem[{Zoph et~al.(2016)Zoph, Yuret, May, and Knight}]{zoph2016transfer}
Barret Zoph, Deniz Yuret, Jonathan May, and Kevin Knight. 2016.
\newblock Transfer learning for low-resource neural machine translation.
\newblock In \emph{EMNLP}.

\end{thebibliography}
\bibliographystyle{acl_natbib}
\end{document}